\documentclass{llncs}
\usepackage{graphicx}
\usepackage{bbding}
\usepackage{booktabs}
\usepackage[tight,footnotesize]{subfigure}
\usepackage{multirow}
\usepackage{longtable}
\usepackage{amsmath}
\usepackage[ruled]{algorithm2e}
\usepackage{amsfonts,dsfont}
\usepackage[colorlinks,linkcolor=blue,anchorcolor=blue,citecolor=blue]{hyperref}
\usepackage{cite}
\begin{document}
\title{MODRL/D-AM: Multiobjective Deep Reinforcement Learning Algorithm Using Decomposition and Attention Model for \mbox{Multiobjective Optimization}}
%
\titlerunning{MODRL/D-AM}
%
\author{Hong Wu \and
Jiahai Wang\inst{ (}\Envelope\inst{)} \and
Zizhen Zhang
}
%
\authorrunning{Wu et al.}
%
\institute{Department of Computer Science, Sun Yat-sen University, Guangzhou 510006, China
\email{wangjiah@mail.sysu.edu.cn}}
%
\maketitle              
\begin{abstract}
Recently, a deep reinforcement learning method is proposed to solve multiobjective optimization problem. In this method, the multiobjective optimization problem is decomposed to a number of single-objective optimization subproblems and all the subproblems are optimized in a collaborative manner. Each subproblem is modeled with a pointer network and the model is trained with reinforcement learning. However, when pointer network extracts the features of an instance, it ignores the underlying structure information of the input nodes. Thus, this paper proposes a multiobjective deep reinforcement learning method using decomposition and attention model to solve multiobjective optimization problem. In our method, each subproblem is solved by an attention model, which can exploit the structure features as well as node features of input nodes. The experiment results on multiobjective travelling salesman problem show the proposed algorithm achieves better performance compared with the previous method.
\end{abstract}
\section{Introduction}
A multiobjective optimization problem (MOP) can be defined as follows:
\begin{equation}
\begin{aligned}
&\mbox{min}\  f(\textbf{x})\ =\  (f_{1}(\textbf{x}),\  f_{2}(\textbf{x}), \dots, f_{m}(\textbf{x})) \\
&\mbox{subject to }\textbf{x} \in S,
\end{aligned}
\end{equation}
where $S$ is the decision space, $f:S\rightarrow \mathbb{R}^m$ is composed of $m$ real-valued objective functions where $\mathbb{R}^m$ is called the objective space, and $f_{i}(\textbf{x})$ for $i \in \{1,2,\dots, m\}$ is the $i$-th objective of the MOP. Since different objectives in the MOP are \mbox{usually} \mbox{conflicting}, it is impossible to find one best solution that can optimize all \mbox{objectives} at the same time. Thus a trade-off is required among different objectives.

Let $u, v \in \mathbb{R}^m$, $u$ is said to dominate $v$ if and only if $u_i\leq v_i$ for every $i \in \{1,2,\dots,m \}$ and $u_j < v_j$ for at least one index $j \in \{1,2,\dots,m \}$. A solution $x^* \in S$ is called a pareto optimal solution if there is no solution $x \in S$ such that $f(x)$ dominates $f(x^*$)~\cite{MOEA/D}. The set of all pareto optimal solutions is named as pareto set (PS), and the set $\{f(s)| s \in \mbox{PS}\}$ is called the pareto front (PF)~\cite{MOEA/D}.

Many \mbox{MOPs} are NP-hard, such as multiobjective \mbox{travelling} salesman problem (MOTSP), multiobjective vehicle routing problem, etc. It is often difficult to find the PF of a MOP using exact algorithms. There are mainly two categories of optimization algorithms for solving MOPs. The first category is heuristics, such as NSGA-$\rm \uppercase\expandafter{\romannumeral2}$~\cite{NSGA} and MOEA/D~\cite{MOEA/D}. The second category is the learning heuristic based methods~\cite{DRL-MOA}. Heuristics are often used to solve MOPs~\cite{WJH1,WJH2,LSDG,PSO}, but there are several drawbacks for them. Firstly, it is time-consuming for heuristics to approximate the PF of a MOP. Secondly, once there is a slight change of the problem, the heuristic may need to re-perform again to compute the solutions~\cite{DRL-MOA}. As a problem-specific method, heuristics often need to be revised for different problems, even for the similar ones.

Recently, some researchers begin to focus on deep reinforcement learning (DRL) for single-objective optimization problem~\cite{RLTSP,TSPPG,GNNCOP,attentionTSP,RLVRP}. Instead of designing specific heuristics, DRL learn heuristics directly from data on end-to-end neural network. \mbox{Taking} travelling salesman problem (TSP) as an example, given $n$ cities as input, the aim is to get a sequence of these cities with minimum tour length. DRL views the problem as a Markov decision problem. Then TSP can be formulated as follows: the state is \mbox{defined} by the features of the partial solution and unvisited cities, the action is \mbox{represented} by the selection of the next city, the reward is the negative path length of a solution, the policy is the heuristic that learning how to make decisions, which is \mbox{parameterized} by a neural network. The aim of DRL is to train the policy that maximizes the reward. Once a policy is trained, the solution can be generated directly from one feed forward pass of the trained neural network. Without \mbox{repeatedly} solving instances from the same distribution, DRL is more efficient and requires much less problem-specific expert knowledge than heuristics.

Inspired by MOEA/D and the DRL methods proposed recently, a deep \mbox{reinforcement} learning multiobjective \mbox{optimization} algorithm (DRL-MOA)~\cite{DRL-MOA} is proposed to learn heuristics for solving MOPs. In the DRL-MOA, \mbox{MOTSP} is decomposed to $M$ single-objective optimization \mbox{subproblems} firstly. Then $M$ modified pointer networks, each of them is similar to the pointer network in~\cite{pointer}, are used to model these subproblems. Finally, these models are trained by REINFORCE algorithm~\cite{REINFORCE} sequentially. The experiment results on MOTSP in~\cite{DRL-MOA} show that DRL-MOA achieves better performance than NSGA-$\rm\uppercase\expandafter{\romannumeral2}$ and MOEA/D.

As we know, MOTSP is defined on a graph, every node in the graph not only contains its own features, but also the graph structure features such as the distances from other nodes. In DRL-MOA, when the modified pointer network models the subproblems of MOTSP, it does not consider the graph structure features of the graph. Therefore, this paper proposes a multiobjective deep reinforcement learning algorithm using decomposition and attention model (MODRL/D-AM) to solve MOPs. Attention model can extract the node features as well as graph structure features of MOP instances, which is helpful in making decisions. To show the effectiveness of our method, MODRL/D-AM is compared with DRL-MOA for solving MOTSP, and a significant improvement is observed in the overall performance of convergence and diversity.

The remainder of our paper is organized as follows. In Section 2, DRL-MOA is described. MODRL/D-AM is introduced in Section 3. Experiment results and analysis are presented in Section 4. Finally, conclusions are given in Section 5.

\section{Brief Review of DRL-MOA for MOTSP}

\subsection{Problem Formulation and Framework}
We focus on MOTSP in this paper. Given $n$ cities and $m$ objective functions, and the $j$-th objective function of MOTSP is formulated as follows~\cite{MOTSP}:
\begin{equation}
\begin{aligned}
f_j(\pi) = \sum^{n-1}_{i=1}c^j_{\pi(i),\pi(i+1)} + c^j_{\pi(n),\pi(1)}, \quad
&j \in \{1, 2, \dots,m \},
\end{aligned}
\end{equation}
where route $\pi$ is a permutation of $n$ cities and $c^j_{\pi(i),\pi(i+1)}$ is the $j$-th cost from city $\pi(i)$ to city $\pi(i+1)$. The goal of MOTSP is to find a set of routes that minimize the $m$ objective functions simultaneously.

Just like MOEA/D, DRL-MOA decomposes MOTSP to $M$ scalar optimization subproblems by the well-known weighted sum approach, which considers the combination of different objectives. Let $\textbf{$\lambda_i$} = (\lambda_{i1}, \lambda_{i2}, \dots, \lambda_{im})$, where $\lambda_{ij} \geq 0, j \in \{1, \dots, m\}$ and $\sum_{j=0}^{m} \lambda_{ij} = 1$, be a weight vector corresponding to the $i$-th scalar optimization subproblem of MOTSP, which is defined as follows:
\begin{equation}
\label{equation3}
g^{ws}(\pi|\lambda_i) = \sum^{m}_{j=1}\lambda_{ij}f_{j}(\pi)
\end{equation}

\begin{figure}[t]
\centering
\subfigure[]{\includegraphics[width=0.42\linewidth]{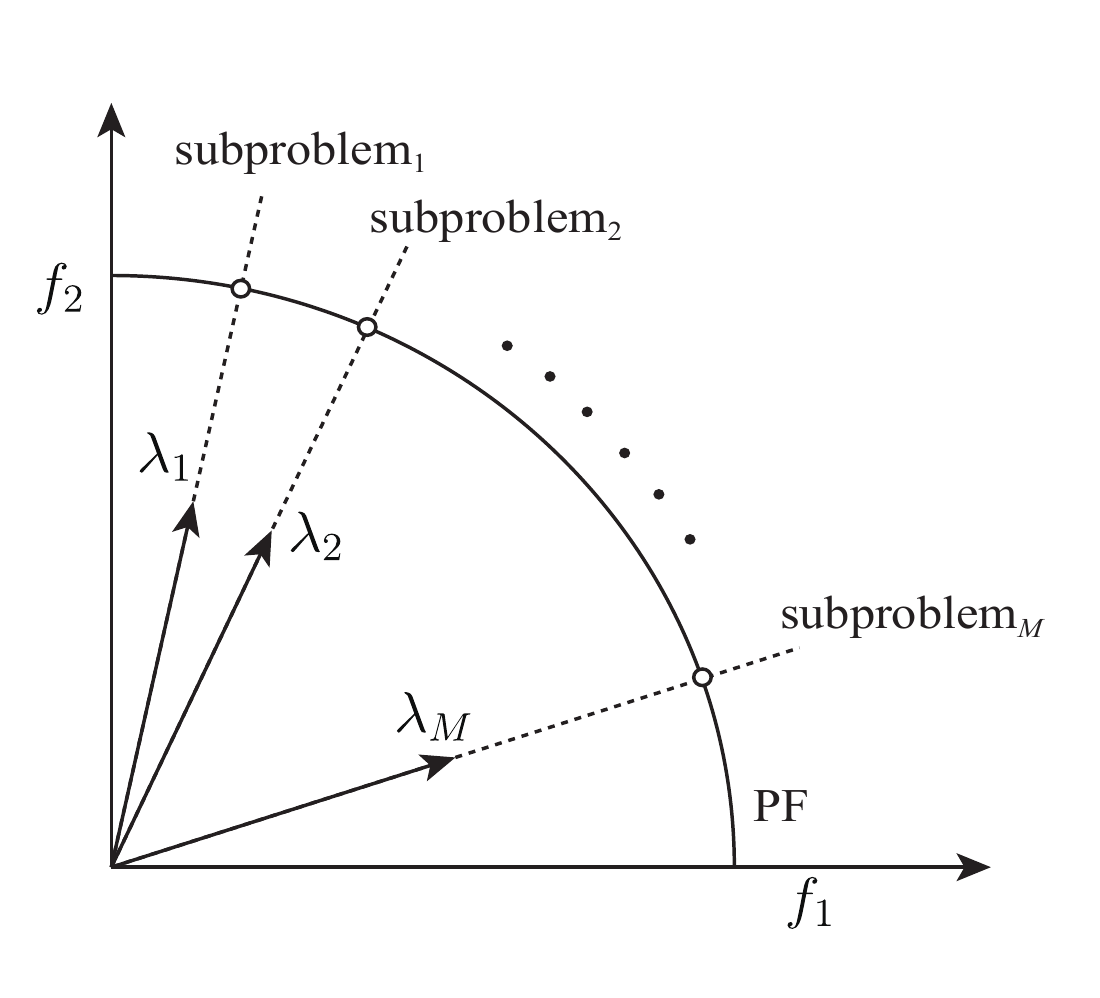}}
\subfigure[]{\includegraphics[width=0.42\linewidth]{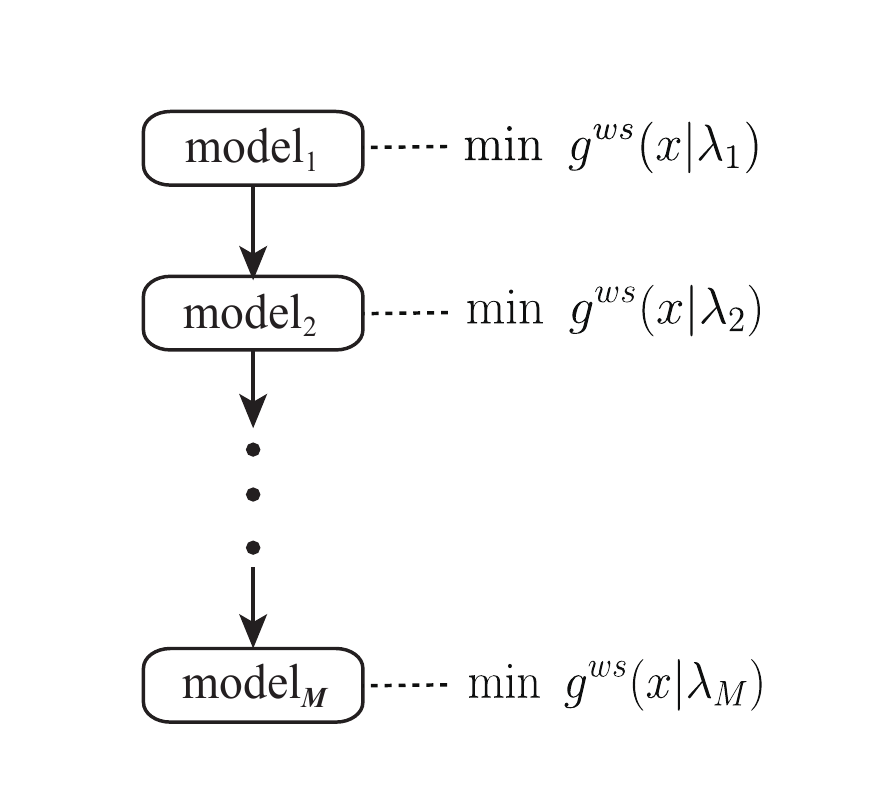}}
\caption{(a) Decomposition strategy. (b) Neighborhood-based transfer strategy.}
\label{figure1}
\end{figure}

The optimal solution of the scalar optimization problem above is a pareto optimal solution. Then, let $\{ \lambda_1, \lambda_2, \dots, \lambda_M \}$ be a set of weight vectors, each weight vector corresponds to a scalar optimization subproblem. When $m = 2$, the weight vectors and corresponding subproblems are spread uniformly as in Fig.~\ref{figure1} (a). The PS is made up of the non-dominated solutions of all subproblems.

After decomposing MOTSP to a set of scalar optimization subproblems, each subproblem can be modelled by a neural network and solved by DRL methods. However, training $M$ models requires huge amount of time. Thus, to decrease the training time of the models, DRL-MOA adopts a neighborhood-based transfer strategy, which shows in Fig.~\ref{figure1} (b). Each model corresponds to a subproblem. When one subproblem is solved, the parameters of the corresponding model will be transferred to the model of the neighborhood subproblem, then the neighborhood subproblem will be solved quickly. By making use of the neighborhood information among subproblems, all subproblems are tackled sequentially in a quick manner. The basic idea of DRL-MOA is shown in Algorithm~\ref{algorithm1}. The subproblems are solved sequentially and $M$ models are trained with REINFORCE algorithm by combining DRL and neighborhood-based transfer strategy. Finally, the PF can be approximated by a simple feed forward calculation of the $M$ models.

\begin{algorithm}[H]
\caption{Framework of DRL-MOA}
\LinesNumbered
\label{algorithm1}
\KwIn{a well spread weight vectors {$\{\lambda_1, \lambda_2, \dots, \lambda_M \}$}, the model of subproblems $\omega$ }
\KwOut{the optimal model $\omega^*$}
$\omega_{\lambda_1} \leftarrow$  initialization\;
\For{$i = 1:M$}{
    \eIf{$i == 1$}{
        $\omega^*_{\lambda_1} \leftarrow$ REINFORCE($\omega_{\lambda_1}, g^{ws}_{\lambda_1}$)\;
    }
    {
        $\omega_{\lambda_i} \leftarrow \omega^*_{\lambda_{i-1}}$\;
        $\omega^*_{\lambda_i} \leftarrow$ REINFORCE($\omega_{\lambda_i}, g^{ws}_{\lambda_i}$)\;
    }
}
\Return $\omega^*$
\end{algorithm}

\subsection{Model of Subproblem: Pointer Network}
A subproblem instance of MOTSP can be defined in a graph with $n$ nodes, which is denoted by a set $X = \{x_1, x_2, \dots, x_n\}$. Each node $x_i$ has a feature vector $(x_{i1}, x_{i2}, \dots, x_{im})$, which corresponds to the $m$ different objectives of MOTSP. For example, a feature used widely is the 2-dimensional coordinate of Euclidean space. The solution denoted by $\pi = (\pi_1, \dots, \pi_n)$ is a permutation of the graph nodes of MOTSP. The objective is minimizing the weighted sum of different objectives like Eq.~(\ref{equation3}). The process of generating a solution can be viewed as a sequential decision process,
so each subproblem can be solved by an encoder-decoder model~\cite{ED} parameterized by $\theta$. Firstly, the encoder maps the node features to node embeddings in a high-dimensional vector space. Then the decoder generates the solution step by step. At each decoding step $t \in \{1, 2, 3, \dots, n\}$, one node $\pi_{t}$ that has not been visited is selected. Hence, the probability of a solution can be modelled by the chain rule:
\begin{equation}
p_{\theta}(\pi|X) = \prod_{t=1}^{n}p_{\theta}(\pi_{t}|\pi_{1:t-1}, X).
\label{eq4}
\end{equation}

In DRL-MOA, a modified pointer network is used to compute the probability in Eq.~(\ref{eq4}). The encoder of the modified pointer network transforms each node feature to an embedding in a high-dimensional vector space through a 1-dimensional (1-D) convolution layer. At each decoding time $t$, a gated recurrent unit (GRU)~\cite{ED} and a variant of attention mechanism~\cite{NMT} are used to produce a probability distribution over the unvisited nodes, which is used to select the next node to visit. More details of the modified pointer network can be found in~\cite{DRL-MOA}.

\section{The Proposed Algorithm: MODRL/D-AM}
\subsection{Motivation}
In DRL-MOA, a modified pointer network is used to model the subproblem of MOTSP. In the modified pointer network, an encoder extracts the node features using a simple 1-D convolutional layer. However, each subproblem of MOTSP is defined over a graph that is fully-connected (with self-connections). Such a simple encoder can not exploit the graph structure of a problem instance. At the decoding time $t$, the decoder uses a GRU to map a partial tour $\pi_{1:t-1}$ to a hidden state, which is used as decoding context to calculate the probability distribution of selecting the next node. However, the partial tour can not be changed and our goal is to construct a path from $\pi_{t-1}$ to $\pi_1$ through all unvisited nodes. In other words, the selection of the next node is relevant only to the first and last node of the partial tour. Using a GRU in modified pointer network to map the total partial path to a hidden state may be not so helpful in selecting the next node, since there is much irrelevant information in the hidden state. Thus, this paper uses the attention model~\cite{attentionTSP}, instead of the pointer network, to model the subproblem.
\subsection{Model of Subproblem: Attention Model}
\begin{figure}[t]
\centering
\subfigure[]{\includegraphics[width=0.35\linewidth]{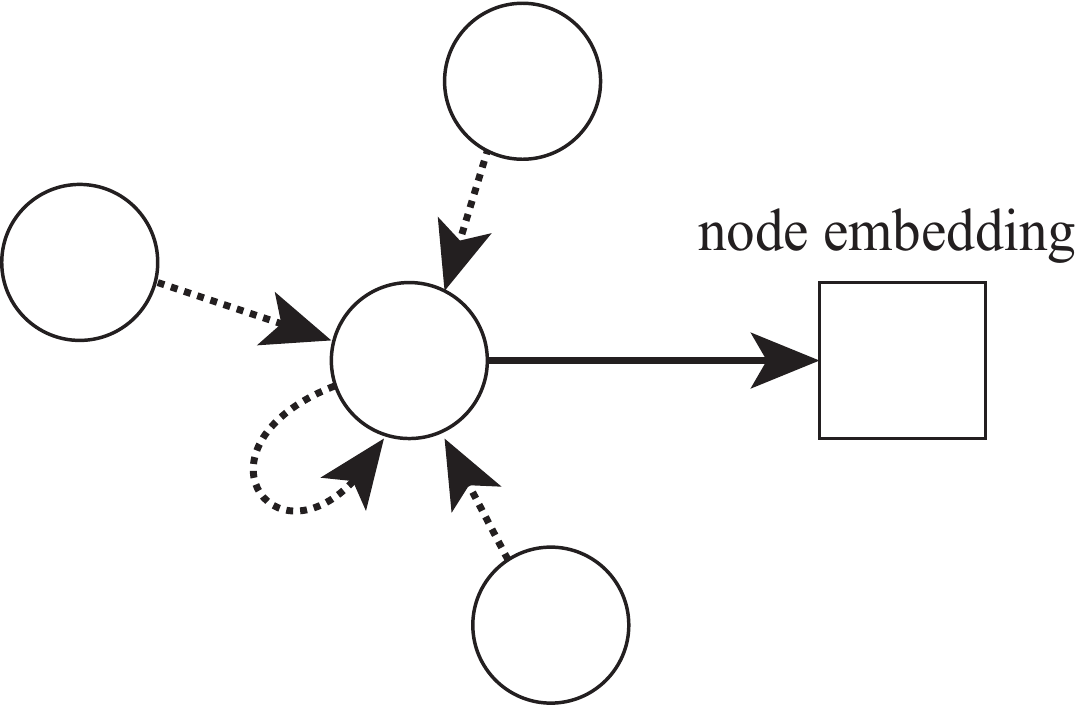}}
\subfigure[]{\includegraphics[width=0.64\linewidth]{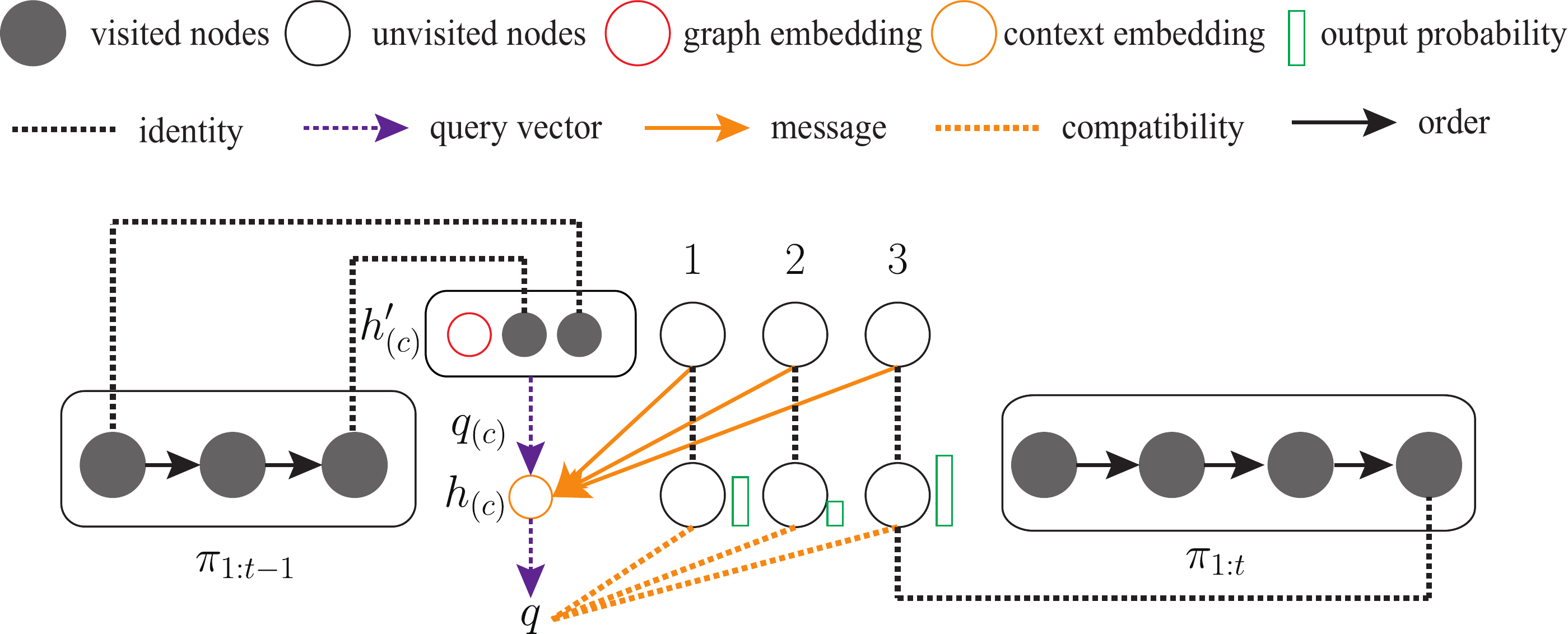}}
\caption{(a) Encoder of attention model. (b) Decoding process at decoding step $t$.}
\label{figure3}
\end{figure}
The attention model is also an encoder-decoder model. However, different from the modified pointer network, the encoder of attention model can be viewed as a graph attention network~\cite{GAN}, which is used to compute the embedding of each node. As show in Fig.~\ref{figure3} (a), by attending over other nodes, the embedding of each node contains the node features as well as the structure features. The decoder of attention model does not use a GRU to summarize the total partial path to a decoding context vector. Instead, the decoding context vector is calculated using the graph embedding, the first and last node embeddings of the partial tour, which is more useful in selecting the next node. The details of attention model are described below.
\subsubsection{Encoder of Attention Model}
The encoder of attention model transforms each node feature vector in the $d_x$-dimensional vector space to a node embedding in the $d_h$-dimensional vector space. The encoder is consisted of a linear transformation layer and $N$ attention layers, which is similar to the encoder used in the Transformer architecture~\cite{attention}. But the encoder of attention model does not use the positional encoding since the input order is not meaningful. For each node $x_i$, where $i \in \{1, \dots, n\}$, the linear transformation layer with parameters $W \in \mathbb{R}^{d_h \times d_x}$ and $b \in \mathbb{R}^{d_h}$ transforms the node feature vector to the initial node embedding $h_i^0$:
\begin{equation}
h_i^0 = Wx_i +b
\end{equation}
Then the node embeddings $\{h_1^0, \dots, h_n^0\}$ are fed into $N$ attention layers. Each attention layer contains a multi-head attention sublayer and a feed-forward sublayer. For each sublayer, a batch normalization~\cite{BN} layer and a skip connection~\cite{SC} layer are used to accelerate the training process.
\paragraph{Multi-Head Attention Sublayer}
For each node $x_i$, this sublayer is used to aggregate different types of message from other nodes in the graph. Let the embedding of each node $x_i$ in layer $l$ be $h_i^l$, where $i \in \{1, \dots, n\}$ and $l \in \{1, \dots, N\}$. The output of multi-head attention sublayer $\hat{h}_i^l$ can be computed as follows:
\begin{equation}
\hat{h}_i^l = \mbox{BN}^{l}(h_i^{l-1} + \mbox{MHA}_i^l(h_1^{l-1}, \dots, h_n^{l-1})),
\end{equation}
where $\mbox{BN}$ is the batch normalization layer and $\mbox{MHA}_i^l(h_1^{l-1}, \dots, h_n^{l-1})$ is the multi-head attention vector that contains different type of messages from other nodes. The number of heads is set to $A$. For each head $a$, the query vector $q_{ia}^l \in \mathbb{R}^{d_k}$, the key vector $k_{ia}^l \in \mathbb{R}^{d_k}$ and the value vector $v_{ia}^l \in \mathbb{R}^{d_v}$ is calculated by a transformation of the node embedding $h_i^{l-1}$ for each node ($d_k = d_v = \frac{d_h}{A}$). Then the process of computing the multi-head attention vector is described as follows:
\begin{equation}
q_{ia}^l = W_{qa}^lh_i^{l-1}, k_{ia}^l = W_{ka}^lh_i^{l-1}, v_{ia}^l = W_{va}^lh_i^{l-1},
\end{equation}
\begin{equation}
u_{ija}^l = \frac{(q_{ia}^l)^{T}k_{ja}^l}{\sqrt{d_k}}, \quad
w_{ija}^l = \frac{e^{u_{ija}^l}}{\sum_{j'=1}^ne^{u_{ij'a}^l}},
\end{equation}
\begin{equation}
h_{ia}^{l} = \sum_{j=1}^nw_{ija}^lv_{ja}^l, \quad
\mbox{MHA}_i^l(h_1^{l-1}, \dots, h_n^{l-1}) = \sum_{a = 1}^AW_{oa}^lh_{ia}^l,
\end{equation}
where $W_{qa}^l \in \mathbb{R}^{d_k \times d_h}, W_{ka}^l \in \mathbb{R}^{d_k \times d_h}, W_{va}^l \in \mathbb{R}^{d_v \times d_h}, W_{oa}^l \in \mathbb{R}^{d_h \times d_v}$ are trainable attention weights of the $l$-th multi-head attention sublayer. $u_{ija}^l \in \mathbb{R}$ is the compatibility of the query vector $q_{ia}^l$ of node $x_i$ with the key vector $k_{ja}^l$ of node $x_j$, the attention weight $w_{ija}^l \in [0, 1]$ is calculated using a softmax function. $h_{ia}^{l}$ is the combination of messages from other nodes received by node $x_i$. The multi-head attention vector is computed with $W_{oa}^l$ and $h_{ia}^{l}$.
\paragraph{Feed Forward Sublayer}In this sublayer, the node embedding of each node is updated by making use of the output of the multi-head attention layer. The feed forward sublayer (FF) is consisted of a fully-connected layer with ReLu activation function and another fully-connected layer. For each node $x_i$, the input of the feed forward sublayer is the output of the multi-head attention layer $\hat{h}_i^l$, the output is calculated as follows:
\begin{equation}
\mbox{FF}^l(\hat{h}_i^l) = W_1^{l}ReLu(W_0^{l}\hat{h}_i^l+b_0^{l}) + b_1^{l},
\end{equation}
\begin{equation}
h_i^l = \mbox{BN}^l(\hat{h}_i^l + \mbox{FF}^l(\hat{h}_i^l)),
\end{equation}
where $W_0^{l} \in \mathbb{R}^{d_f \times d_h}, W_1^{l} \in \mathbb{R}^{d_h \times d_f}, b_0^l \in \mathbb{R}^{d_f}$ and $b_1^l \in \mathbb{R}^{d_h}$ are trainable parameters.

For each node $x_i$, the final node embedding $h_i^N$ is calculated by $N$ attention layers. Besides that, the graph embedding $\bar{h}^N$ is defined as follows:
\begin{equation}
\bar{h}^N = \frac{1}{n}\sum_{i=1}^nh_i^N,
\end{equation}
both of the node embeddings and graph embedding will be passed to the decoder.

\subsubsection{Decoder of Attention Model}
At each decoding step $t \in \{1, \dots, n\}$, the decoder needs to make a decision of $\pi_t$ based on the partial tour $\pi_{1:t-1}$, the embeddings of each node and the total graph. Firstly, the initial context embedding $h_{(c)}' \in \mathbb{R}^{3d_h}$ is calculated by a concatenation of the graph embedding $\bar{h}^N$, the node embedding of the first node $h_{\pi_1}^N$ and the last node $h_{\pi_{t-1}}^N$. When $t=1$, $h_{\pi_{1}}^N, h_{\pi_{t-1}}^N$ are replaced by two trainable parameter vectors $v^1 \in \mathbb{R}^{d_h}, v^f \in \mathbb{R}^{d_h}$:
\begin{equation}
h_{(c)}' = \begin{cases}[\bar{h}^N, h_{\pi_1}^N, h_{\pi_{t-1}}^N] \quad t > 1\\ [\bar{h}^N, v^1, v^f] \quad \quad \ \  t = 1.\end{cases}
\end{equation}

Then a new context embedding $h_{(c)}$ is computed with an $A$-head attention layer. The query vector $q_{(c)a} \in \mathbb{R}^{d_k}$ comes from the previous context embedding $h_{(c)}'$. For each node $x_i$, the key vector $k_{ia}^{N+1}\in \mathbb{R}^{d_k}$ and the value vector $v_{ia}^{N+1}\in \mathbb{R}^{d_v}$ are transformed from the node embedding $h_i^N$:
\begin{equation}
q_{(c)a} = W_{qa}'h_{(c)}', k_{ia}^{N+1} = W_{ka}^{N+1}h_i^N, v_{ia}^{N+1} = W_{va}^{N+1}h_i^N,
\end{equation}
where $W_{qa}' \in \mathbb{R}^{d_k \times 3d_h}, W_{ka}^{N+1} \in \mathbb{R}^{d_k \times d_h}$ and $W_{va}^{N+1} \in \mathbb{R}^{d_v \times d_h}$. Then the compatibilities of the query vector with all nodes are computed. Different from the encoder of attention model, the nodes that have been visited are masked when calculating the compatibilities:
\begin{equation}
u_{(c)ia} = \begin{cases}\frac{q_{(c)a}^Tk_{ia}^{N+1}}{\sqrt{d_k}} \quad \quad \quad x_i \notin \pi_{1:t-1}\\ -\infty \quad \quad \quad \quad \quad  \mbox{otherwise.}\end{cases}
\end{equation}

Then the attention weights can be obtained by a softmax function and the new context embedding $h_{(c)}$ can be calculated as follows:
\begin{equation}
w_{(c)ia} = \frac{e^{u_{(c)ia}}}{\sum_{i'=1}^{n}e^{u_{(c)i'a}}},
\end{equation}
\begin{equation}
h_{(c)a} = \sum_{i=1}^nw_{(c)ia}v_{ia}^{N+1}, \quad
h_{(c)} = \sum_{a = 1}^AW_{oa}^{N+1}h_{(c)a},
\end{equation}
where $W_{oa}^{N+1} \in \mathbb{R}^{d_h \times d_v}$. Finally, based on the new context embedding $h_{(c)}$, the probability of selecting node $x_i$ as the next node to visit $p_\theta(\pi_t = x_i|\pi_{1:t-1}, X)$ is calculated by a single-head attention layer:
\begin{equation}
q = W_qh_{(c)}, \quad k_i = W_kh_i^N,
\end{equation}
\begin{equation}
\label{equation23}
u_{i} = \begin{cases}C\cdot \mbox{tanh}(q^{T}k_i) \quad \quad \quad x_i \notin \pi_{1:t-1}\\ -\infty \quad \quad \quad \quad \quad \quad \quad \ \   \mbox{otherwise,}\end{cases}
\end{equation}
\begin{equation}
p_\theta(\pi_t = x_i|\pi_{1:t-1}, X) = \frac{e^{u_{i}}}{\sum_{i'=1}^{n}e^{u_{i'}}},
\end{equation}
where $W_q \in \mathbb{R}^{d_h \times d_h}$ and $W_k \in \mathbb{R}^{d_h \times d_h}$ are trainable parameters. When we compute the compatibilities in Eq.~(\ref{equation23}), the result are limited in $[-C, C]$ ($C=10$) by a tanh function.

The decoding process at decoding step $t$ is shown in Fig.~\ref{figure3} (b). Firstly, the context embedding is computed with a multi-head attention layer by making use of the partial solution and unvisited nodes. Then based on the context embedding and unvisited nodes, the probability distribution over unvisited nodes can be calculated by a single-head attention mechanism.

\subsection{Framework and Training Method}
The proposed algorithm uses the same MOEA/D framework as in DRL-MOA (Algorithm~\ref{algorithm1}). The training method is briefly described as follows.

The REINFORCE algorithm, a well-know actor-critic training method, is used to train the model of the subproblem. For each subproblem, the training parameters $\omega_{\lambda_i}$ is composed of an actor network and a critic network. The actor network is the attention model, which is parameterized by $\theta$. The critic network parameterized by $\phi$ has four 1-D convolutional layers to map the embeddings of a problem instance into a single value. The output of the critic network predicts an estimation of the objective function of the subproblem.

For the actor network, the training objective is the weighted sum of different objectives of solution $\pi$ of a problem instance $X$. So the gradients of parameters $\theta$ can be defined as follows:
\begin{equation}
\nabla J(\theta|X)=E_{\pi \sim p_{\theta}(\cdot|X)}[(g^{ws}(\pi|\lambda_i; X)-b_{\phi}(X))\nabla_{\theta}logp_{\theta}(\pi|X)],
\end{equation}
where $g^{ws}(\pi|\lambda_i; X)$ is the objective function of the $i$-th subproblem, which is the weighted sum of different objectives. $\lambda_i$ is the corresponding weight vector. $b_{\phi}(X)$ is a baseline function calculated by the critic network, which estimates the expected objective value to reduce the variance of the gradients.

In the training process, the MOTSP instances are generated from distributions $(\Phi_1, \dots, \Phi_m)$.  Since for each node $x_i$ of an instance $X$, different features $(x_{i1}, \dots, x_{im})$ may come from different distributions $(\Phi_1, \dots, \Phi_m)$. For example, $x_{ij}$ can be a two-dimensional coordinate in Euclidean space and $\Phi_j$ can be a uniform distribution of $[0, 1] \times [0, 1]$. Then the gradients of parameters $\theta$ can be approximated by Monte Carlo sampling as follows:
\begin{equation}
\nabla J(\theta|X) \approx \frac{1}{B} \sum_{j=1}^{B}[(g^{ws}(\pi_j|\lambda_i; X_j)-b_\phi(X_j))\nabla_{\theta}logp_{\theta}(\pi_j|X_j)],
\end{equation}
where $B$ is the batch size, $X_j$ is a problem instance sampled from $(\Phi_1, \dots, \Phi_m)$ and $\pi_j$ generated by the actor network is the solution of $X_j$.

Different from the actor network, the critic network aims to learn to estimate the expected objective value given an instance $X$. Hence, the objective function of the critic network can be a mean squared error function between the estimated objective value of the critic network $b_\phi(X)$ and the actual objective value of the solution generated by the actor network. The objective function of the critic network is formulated as follows:
\begin{equation}
\mathcal{L}_{\phi} = \frac{1}{B} \sum_{j = 1}^{B}(b_{\phi}(X_j)-g^{ws}(\pi_j|\lambda_i; X_j))^2.
\end{equation}

The training algorithm can be described in Algorithm~\ref{algorithm2}.

\begin{algorithm}[H]
\caption{REINFORCE Algorithm}
\LinesNumbered
\label{algorithm2}
\KwIn{batch size $B$, dataset size $D$, number of epochs $E$, the parameters of actor network $\theta$ and the critic network $\phi$}
\KwOut{the optimal parameters $\theta, \phi$}
$\theta, \phi \leftarrow$  initialization from the parameters given in Algorithm~\ref{algorithm1} \\
$T \leftarrow D/B$ \\
\For{$epoch = 1:E$}{
    \For{$t = 1:T$}{
        \For{$j = 1:B$}{
            $X_j \leftarrow$ SampleInstance($\Phi_1, \dots, \Phi_m$)\\
            $\pi_j \leftarrow$ SampleSolution($p_{\theta}(\cdot|X_j)$)\\
            $b_j \leftarrow$ $b_{\phi}(X_j)$
        }
        $d\theta \leftarrow \frac{1}{B} \sum_{j=1}^{B}[(g^{ws}(\pi_j|\lambda_i; X_j)-b_j)\nabla_{\theta}logp_{\theta}(\pi_j|X_j)]$ \\
        $\mathcal{L}_{\phi} \leftarrow \frac{1}{B} \sum_{j = 1}^{B}(b_j-g^{ws}(\pi_j|\lambda_i; X_j))^2$ \\
        $\theta \leftarrow$ ADAM($\theta, d\theta$) \\
        $\phi \leftarrow$ ADAM($\phi, \nabla_{\phi}\mathcal{L}_{\phi}$)
    }
}
\Return $\theta, \phi$
\end{algorithm}

\section{Experiment}

\subsection{Problem Instances and Experimental Settings}
MODRL/D-AM is tested on the Euclidean instances in~\cite{DRL-MOA}. In the Euclidean instances, the two node features are both sampled from $[0, 1] \times [0, 1]$ and both of the two cost functions between node $i$ and node $j$ are the Euclidean distance between them.

To train the models of MODRL/D-AM, problem instances with 20 and 40 nodes are used. After training, two models of MODRL/D-AM are obtained and the influence of different nodes in the training process can be discussed. To show the robustness of our method, the models are tested on problem instances with 20, 40, 100, 150 and 200 nodes. Besides, kroAB100, kroAB150 and kroAB200 generated from TSPLIB~\cite{TSPLIB} are used to test the performance of our method.

DRL-MOA is implemented and used as the baseline. Both our method and DRL-MOA are trained on datasets with 20 and 40 nodes, so there are four models in total: MODRL/D-AM (20), DRL-MOA (20), MODRL/D-AM (40), DRL-MOA (40). To make the result comparison more convincing, some parameters of our method and the baseline are set to the same value. The number of subproblems $M$ is set to 100, the input dimension $d_x$ is set to 4, the dimension of node embedding $d_h$ is set to 128. In the training process, the batch size $B$ is set to 200, the size of problem instances $D$ is set to 500000, and the model of the first subproblem is trained for 5 epochs and each model of the remaining subproblems is trained for 1 epoch. Besides these parameters, the critic network is consisted of four 1-D convolutional layers. The input channels and output channels of the four convolutional layers are (4, 128), (128, 20), (20, 20) and (20, 1), where the first element of a tuple represents the input channel and the second element represents the output channel. For all convolutional layers, the kernel size and stride are set to 1.

In MODRL/D-AM, the number of attention layers $N$ is set to 1, the number of heads $A$ is set to 8, the dimension of the query vector $d_k$ and the value vector $d_v$ are both set to $\frac{d_h}{A}$ = 16, and another dimension in the feed forward sublayer $d_f$ is set to 512.

\subsection{Results and Discussions}
Hypervolume (HV) indicator is calculated to compare the performance of our method and DRL-MOA on tested instances. When computing the HV value, the objective values are normalized and the reference point is set to $(1.2, 1.2)$. The PFs obtained by MODRL/D-AM and DRL-MOA are also compared.
Besides, the influence of different number of nodes in training process is also discussed. All test experiments are conducted by a GPU (GeForce RTX 2080Ti).

\begin{table}[h]
  \centering
  \caption{The average of HV values and the calculation time obtained by MODRL-AM training with 20 and 40 nodes, and DRL-MOA training with 20 and 40 nodes. The test instances are random instances with 20, 40, 70, 100, 150, 200 nodes. The higher HV value is indicated in bold face.}
  \resizebox{1\columnwidth}{!}{
    \begin{tabular}{|c|c|c c|c c|}
    \hline
    \#nodes &       & MODRL/D-AM (20) & DRL-MOA (20) & MODRL/D-AM (40) & DRL-MOA (40) \\
    \hline
    \multirow{2}[4]{*}{20} & HV    & \textbf{0.802} & 0.796 & \textbf{0.796} & 0.785 \\
          & T(s) & 4.6   & 2     & 4.6   & 1.9 \\
    \hline
    \multirow{2}[4]{*}{40} & HV    & \textbf{0.813} & 0.773 & \textbf{0.821} & 0.815 \\
          & T(s) & 8.8   & 4.2   & 8.7   & 4.2 \\
    \hline
    \multirow{2}[4]{*}{70} & HV    & \textbf{0.834} & 0.803 & \textbf{0.856} & 0.842 \\
          & T(s) & 15    & 6.7   & 14.7  & 6.5 \\
    \hline
    \multirow{2}[4]{*}{100} & HV    & \textbf{0.846} & 0.818 & \textbf{0.872} & 0.853 \\
          & T(s) & 21.4  & 9.6   & 20.5  & 10.3 \\
    \hline
    \multirow{2}[4]{*}{150} & HV    & \textbf{0.857} & 0.838 & \textbf{0.884} & 0.864 \\
          & T(s) & 29.5  & 15.9  & 30.7  & 15 \\
    \hline
    \multirow{2}[4]{*}{200} & HV    & \textbf{0.866} & 0.853 & \textbf{0.894} & 0.878 \\
          & T(s) & 39.8  & 18.6  & 38.6  & 20.3 \\
    \hline
    \end{tabular}}%
  \label{table1}%
\end{table}%

\begin{table}[]
  \centering
  \caption{The HV values obtained by MODRL-AM training with 20 and 40 nodes, and DRL-MOA training with 20 and 40 nodes. The test instances are kroAB100, kroAB150, kroAB200. The higher HV value is indicated in bold face.}
  \resizebox{1\columnwidth}{!}{
    \begin{tabular}{|c|c|c c|c c|}
    \hline
    \#nodes &       & MODRL/D-AM (20) & DRL-MOA (20) & MODRL/D-AM (40) & DRL-MOA (40) \\
    \hline
    \multirow{2}[4]{*}{kroAB100} & HV    & \textbf{0.852} & 0.832 & \textbf{0.876} & \textbf{0.876} \\
          & T(s) & 19.8  & 9.2   & 19.2  & 9 \\
    \hline
    \multirow{2}[4]{*}{kroAB150} & HV    & \textbf{0.874} & 0.855 & \textbf{0.891} & 0.884 \\
          & T(s) & 27.9  & 13.7  & 27.4  & 13.5 \\
    \hline
    \multirow{2}[4]{*}{kroAB200} & HV    & \textbf{0.87}  & 0.85  & \textbf{0.885} & 0.882 \\
          & T(s) & 37.3  & 18    & 38    & 17.7 \\
    \hline
    \end{tabular}}%
  \label{table2}%
\end{table}%

The HV values of random instances are shown in Table~\ref{table1}. For the random instances with 20, 40, 70, 100, 150 and 200 nodes, 10 instances are tested for each kind of random instances. The average of the HV values of each kind of random instances is calculated. In terms of the average of HV values, MODRL/D-AM (40) performs better than DRL-MOA (40) in all kinds of random instances. For kroAB100, kroAB150 and kroAB200, the HV values are computed in Table~\ref{table2} and MODRL/D-AM (40) achieves a better performance than DRL-MOA (40). The calculation time of our method is longer than that of DRL-MOA. It is reasonable because the graph attention encoder of attention model requires more calculation resources than a single convolutional layer.

\begin{figure}
\centering
\subfigure[]{\includegraphics[width=0.32\linewidth]{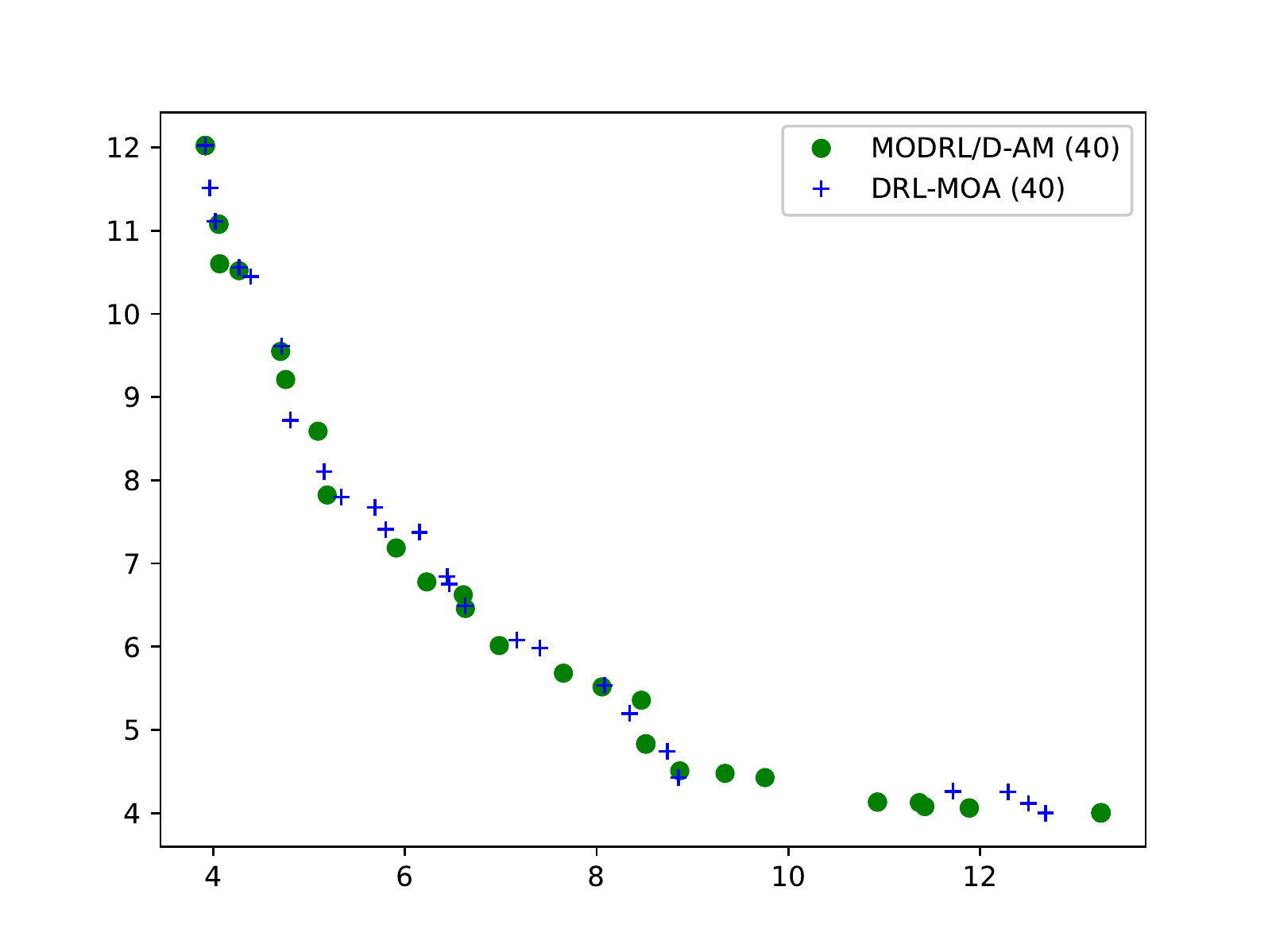}}
\subfigure[]{\includegraphics[width=0.32\linewidth]{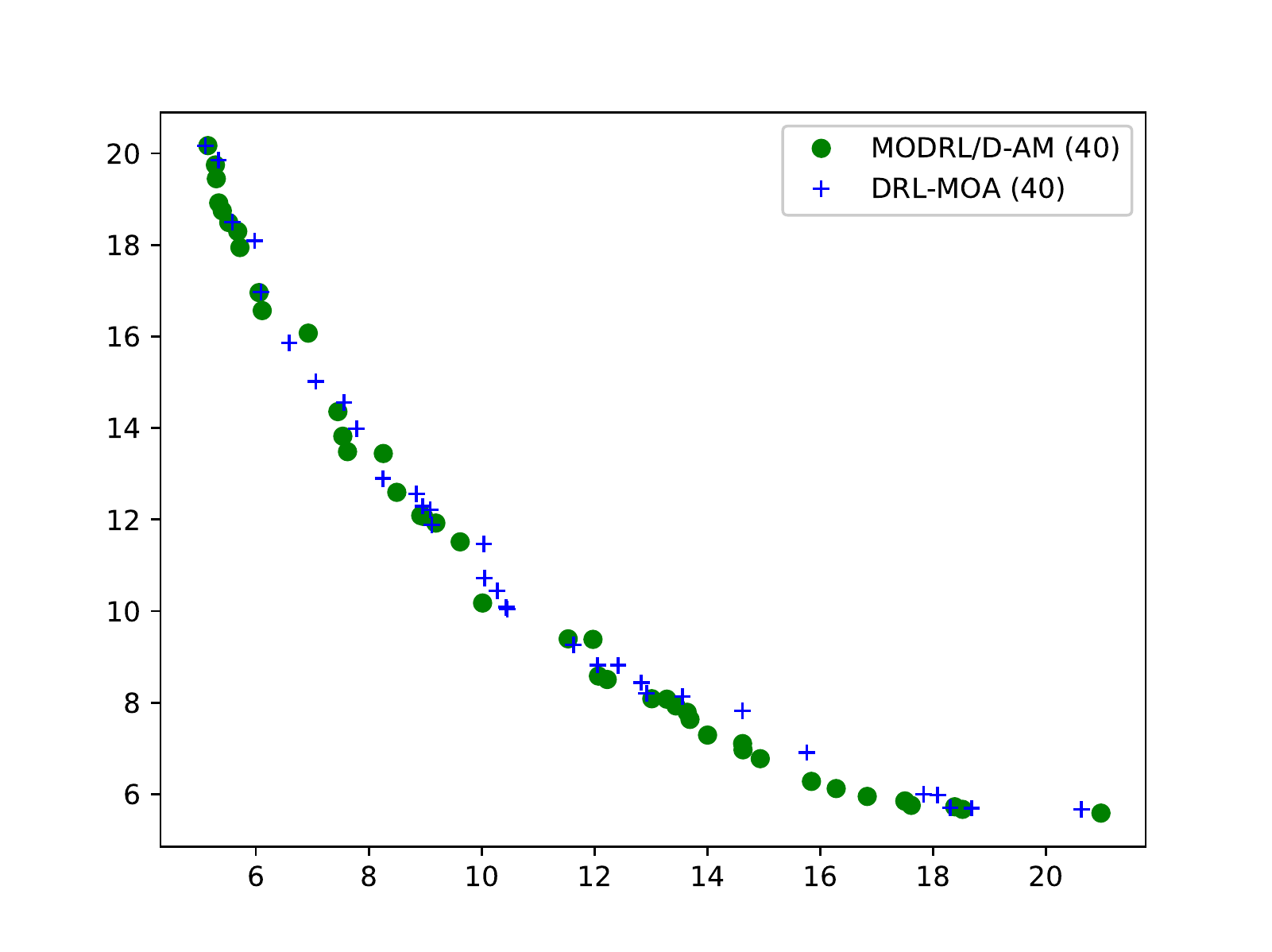}}
\subfigure[]{\includegraphics[width=0.32\linewidth]{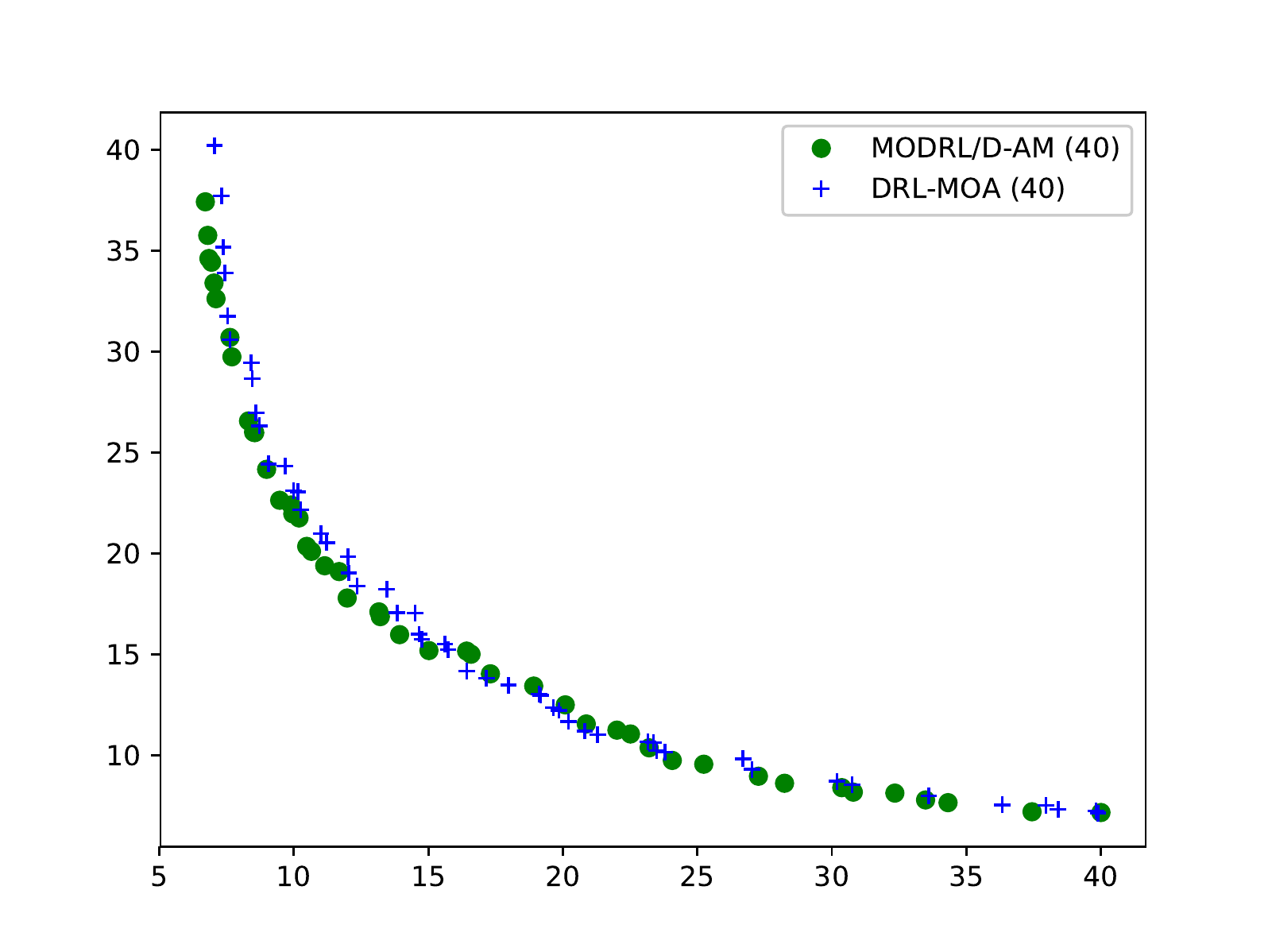}}
\subfigure[]{\includegraphics[width=0.32\linewidth]{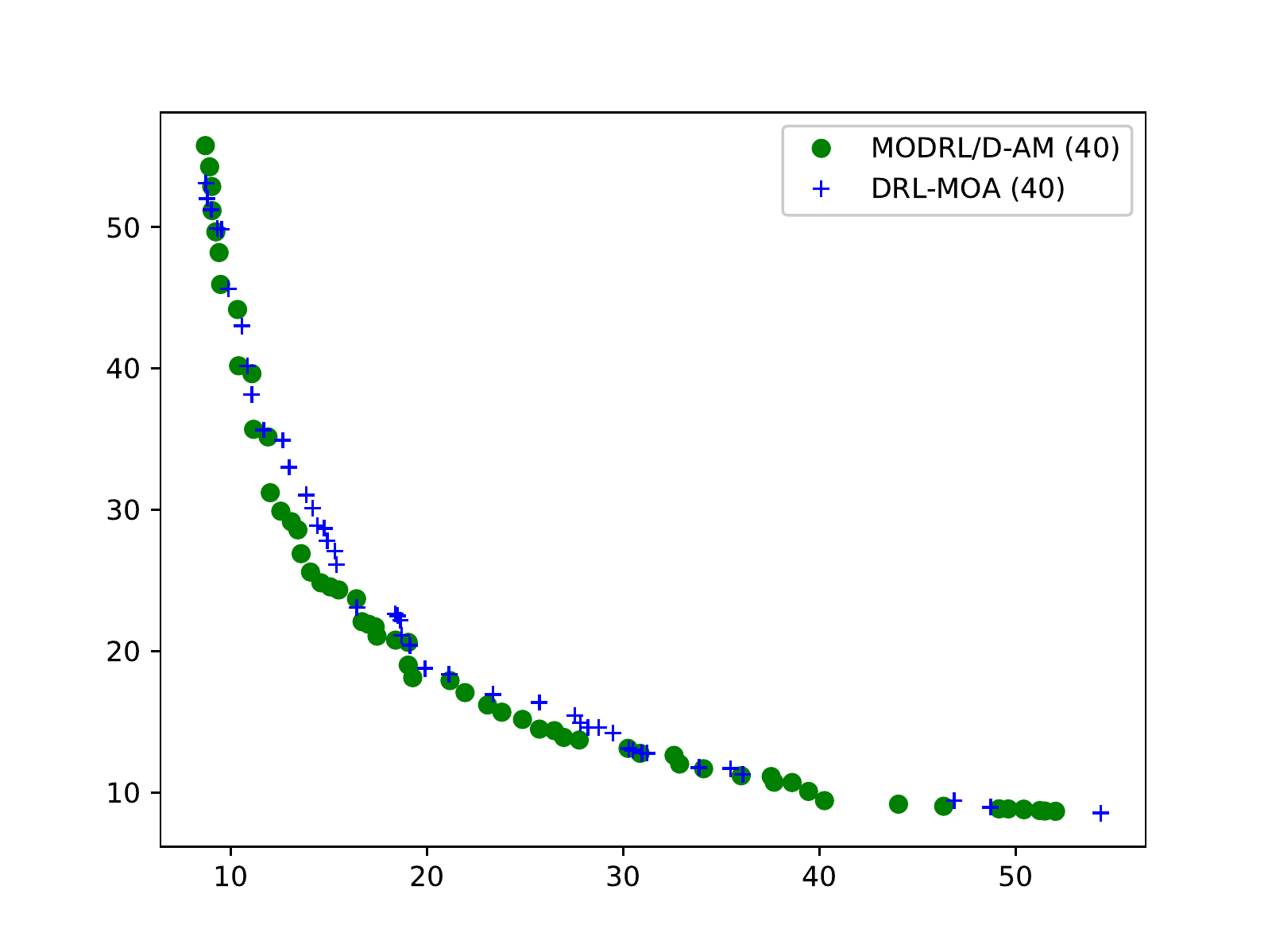}}
\subfigure[]{\includegraphics[width=0.32\linewidth]{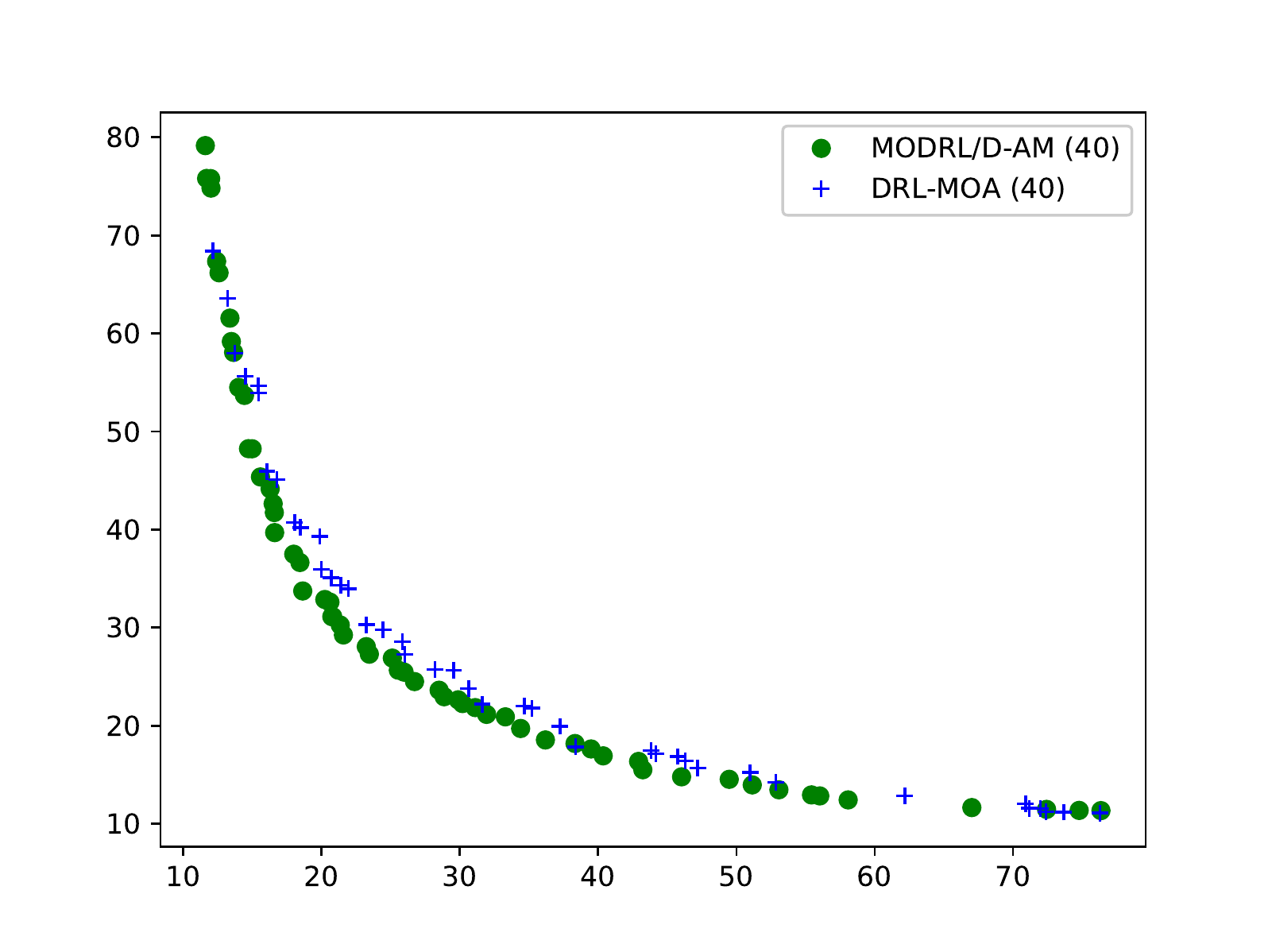}}
\subfigure[]{\includegraphics[width=0.32\linewidth]{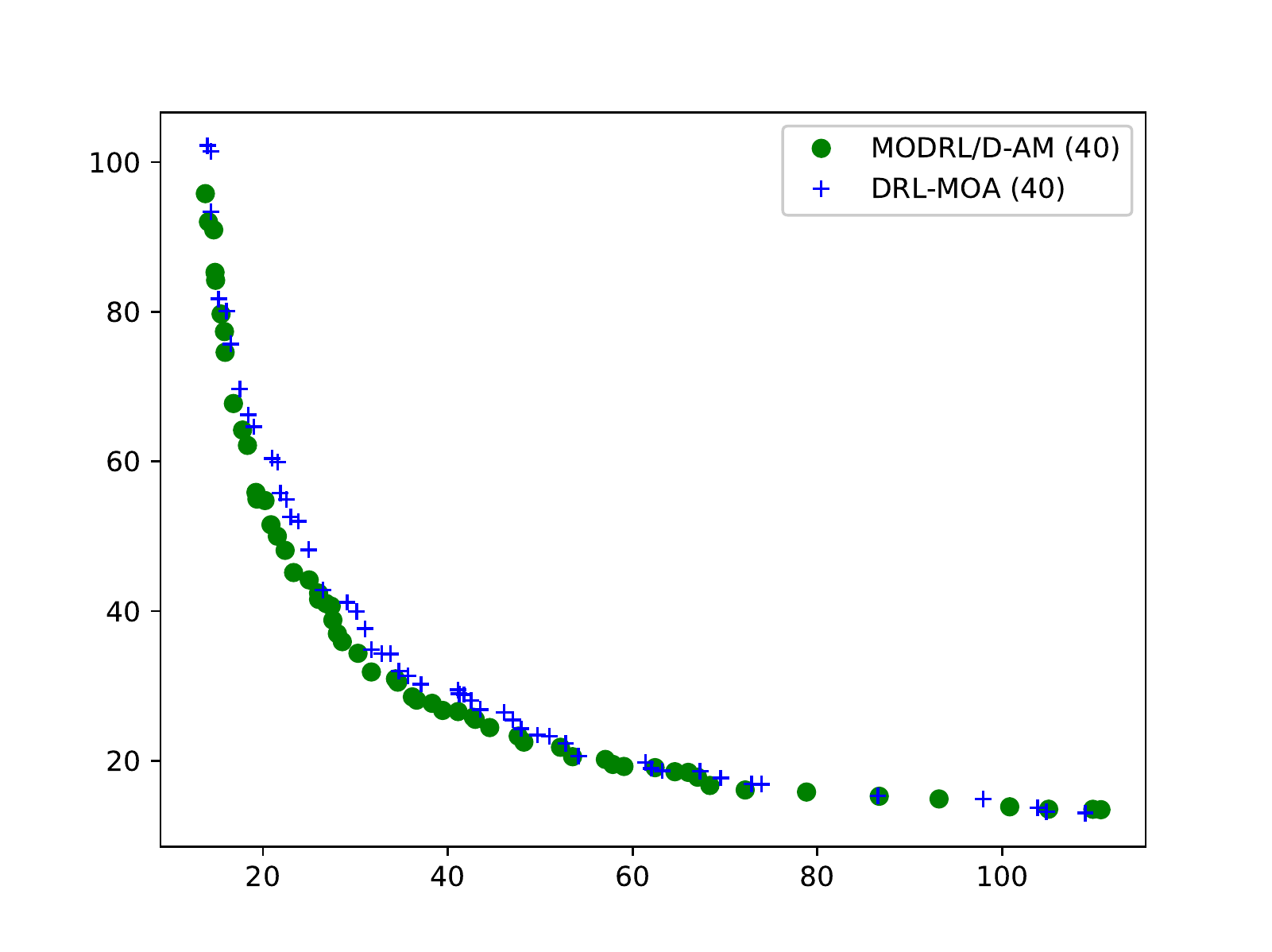}}
\caption{The PFs obtained by MODRL/D-AM (40) and DRL-MOA (40) in solving random instances with (a) 20 nodes, (b) 40 nodes, (c) 70 nodes, (d) 100 nodes, (e) 150 nodes, (f) 200 nodes}
\label{figure4}
\end{figure}
\begin{figure}[t]
\centering
\subfigure[]{\includegraphics[width=0.32\linewidth]{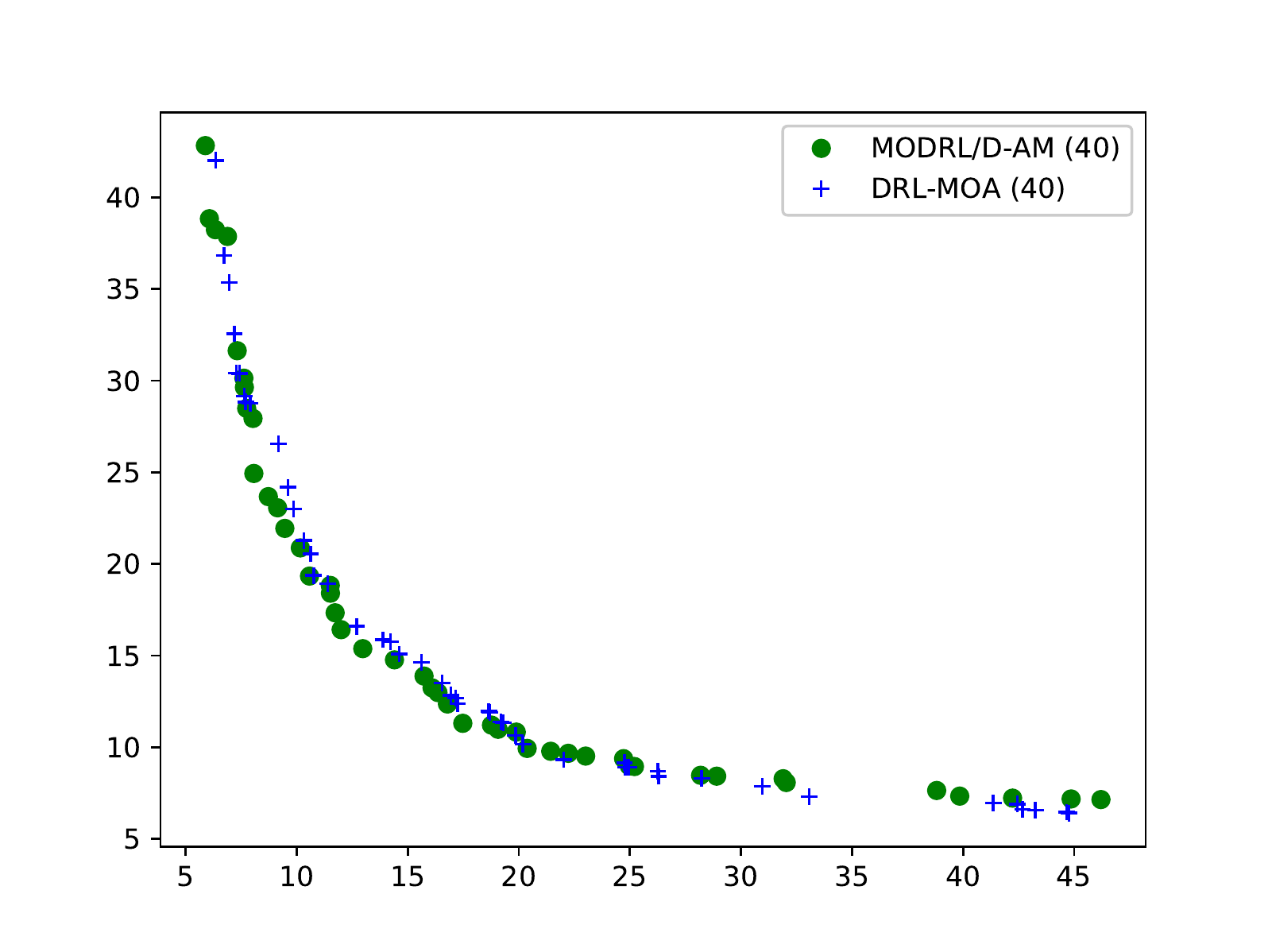}}
\subfigure[]{\includegraphics[width=0.32\linewidth]{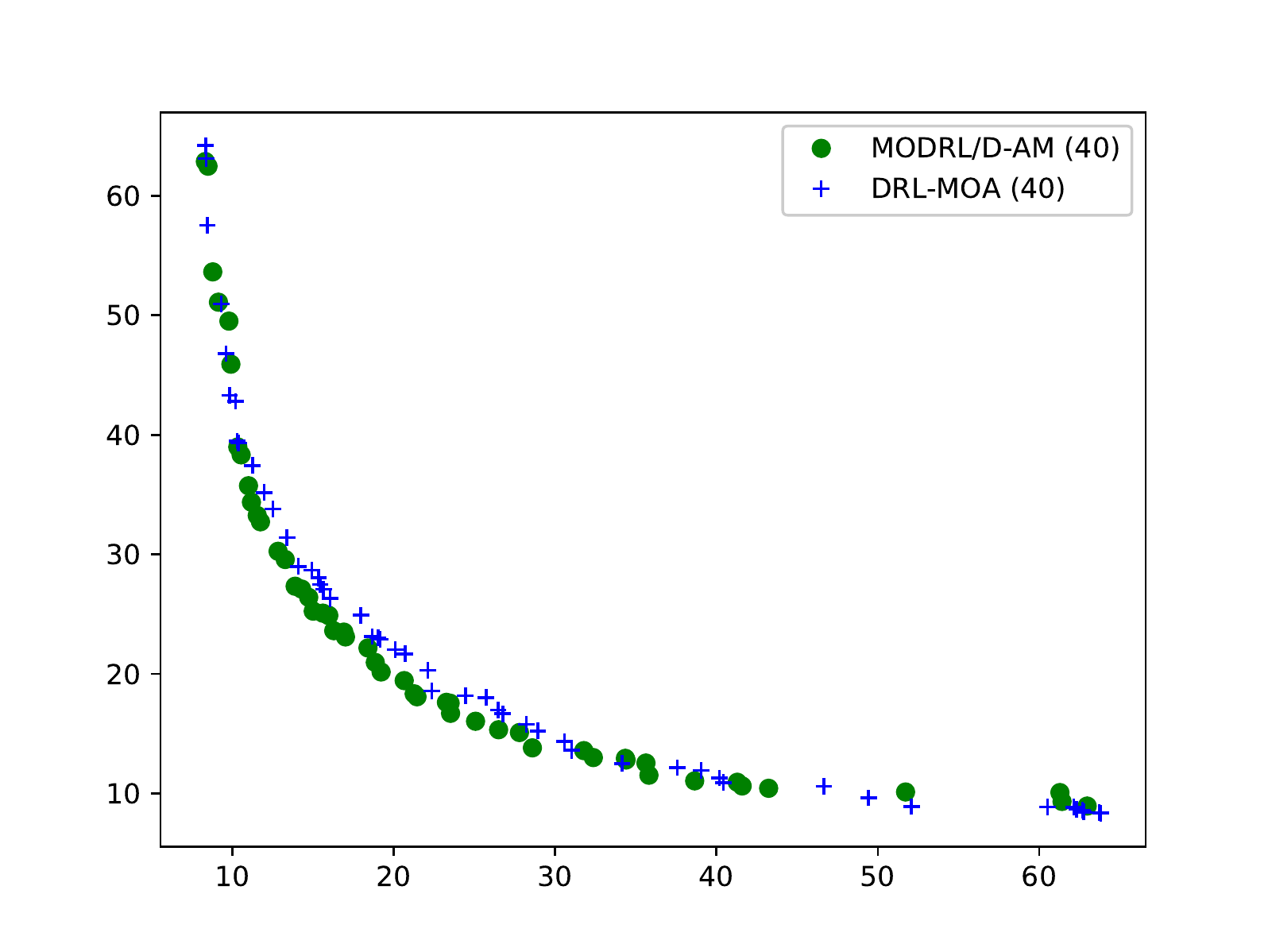}}
\subfigure[]{\includegraphics[width=0.32\linewidth]{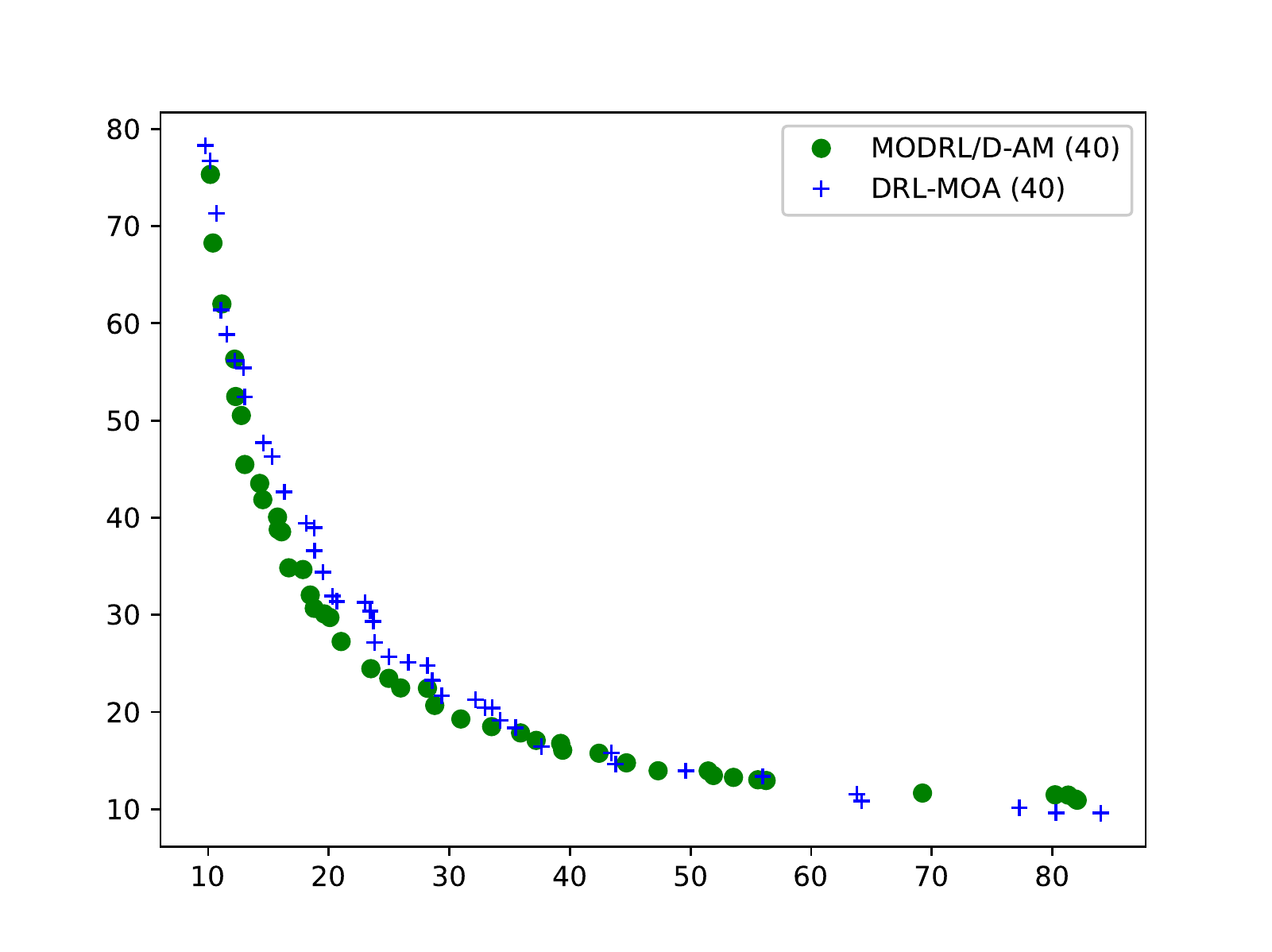}}
\caption{The PFs obtained by MODRL/D-AM (40) and DRL-MOA (40) in solving instances of (a) kroAB100, (b) kroAB150, (c) kroAB200}
\label{figure5}
\end{figure}

\begin{figure}[t]
\centering
\subfigure[]{\includegraphics[width=0.32\linewidth]{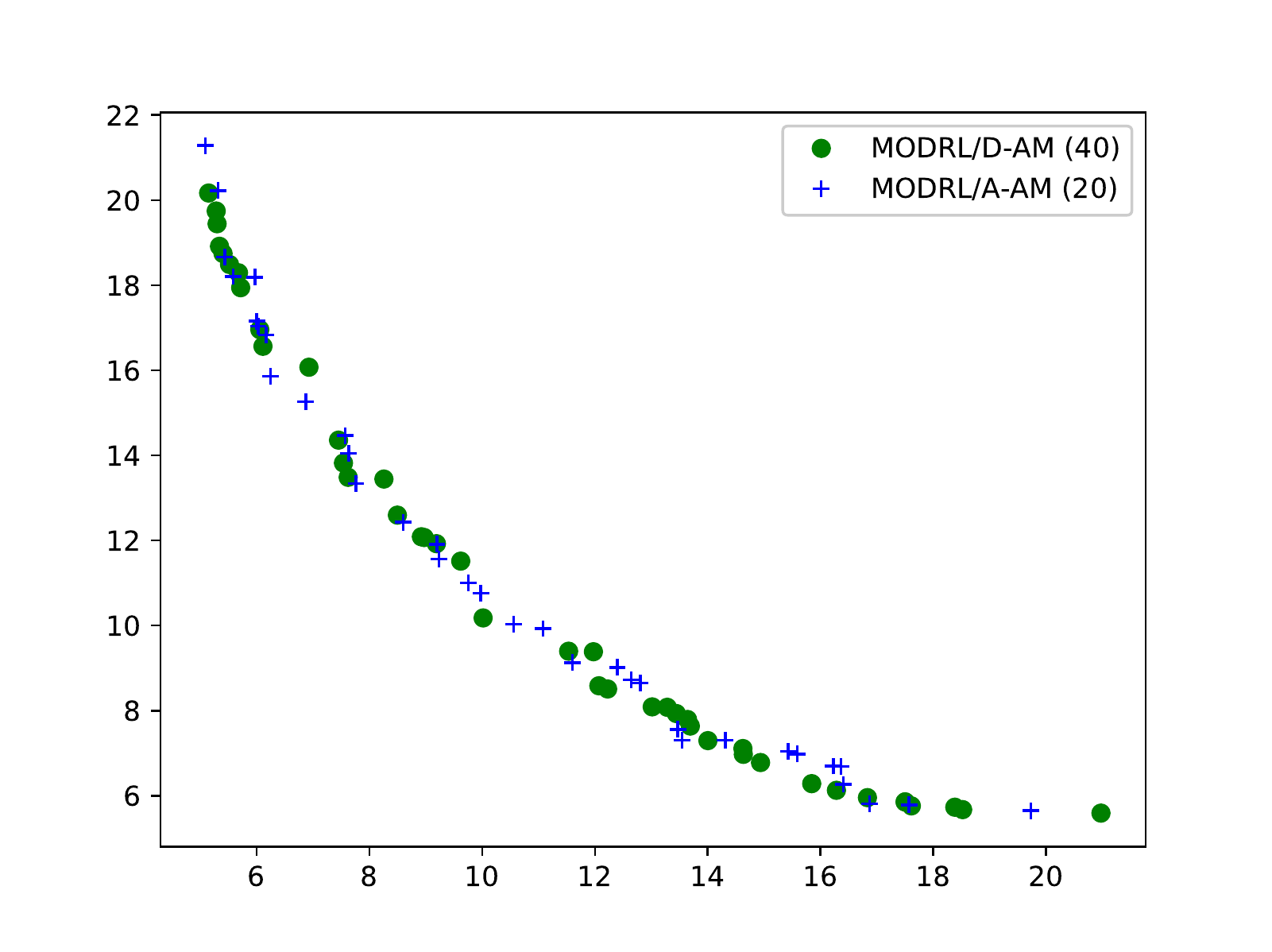}}
\subfigure[]{\includegraphics[width=0.32\linewidth]{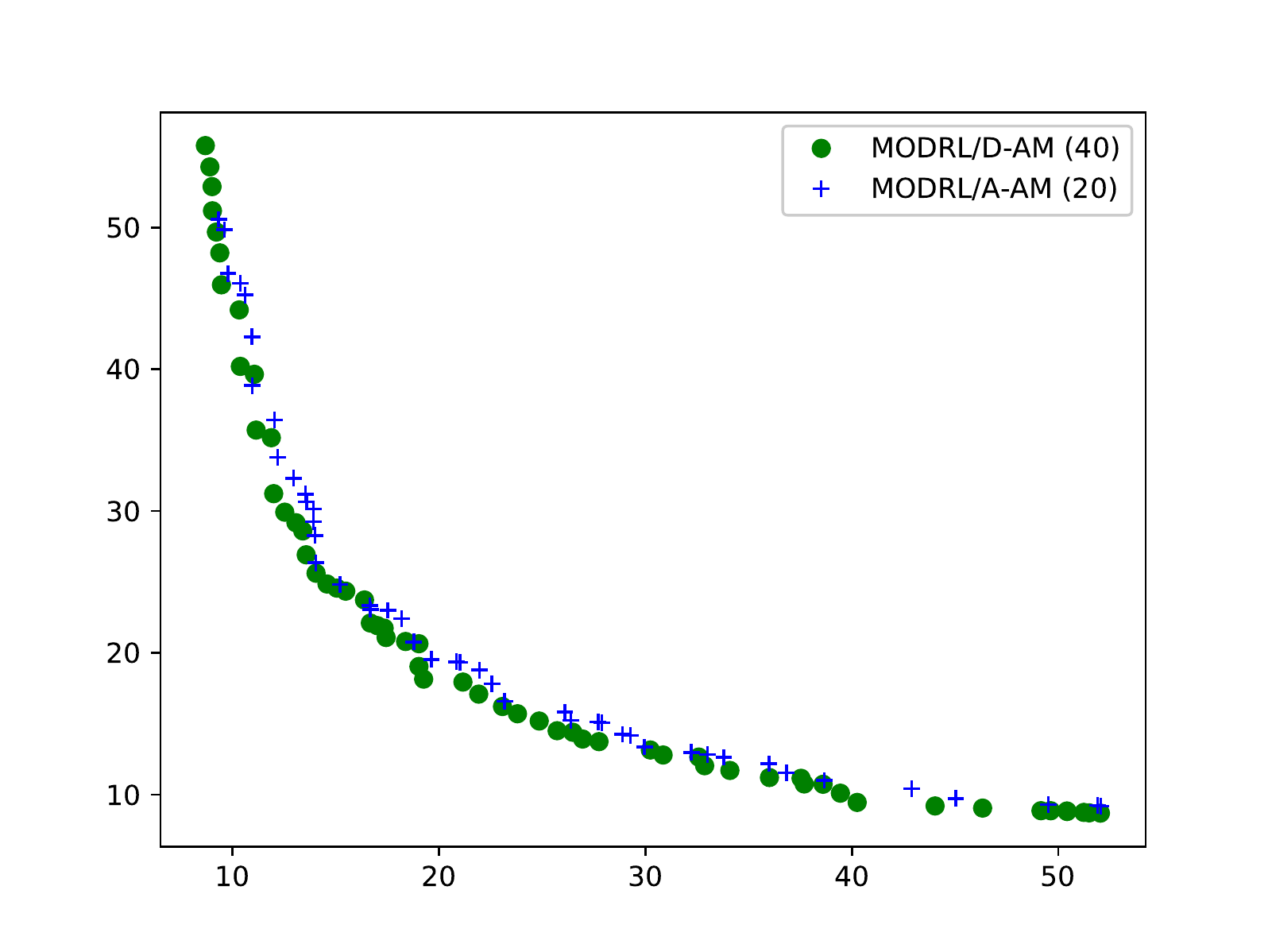}}
\subfigure[]{\includegraphics[width=0.32\linewidth]{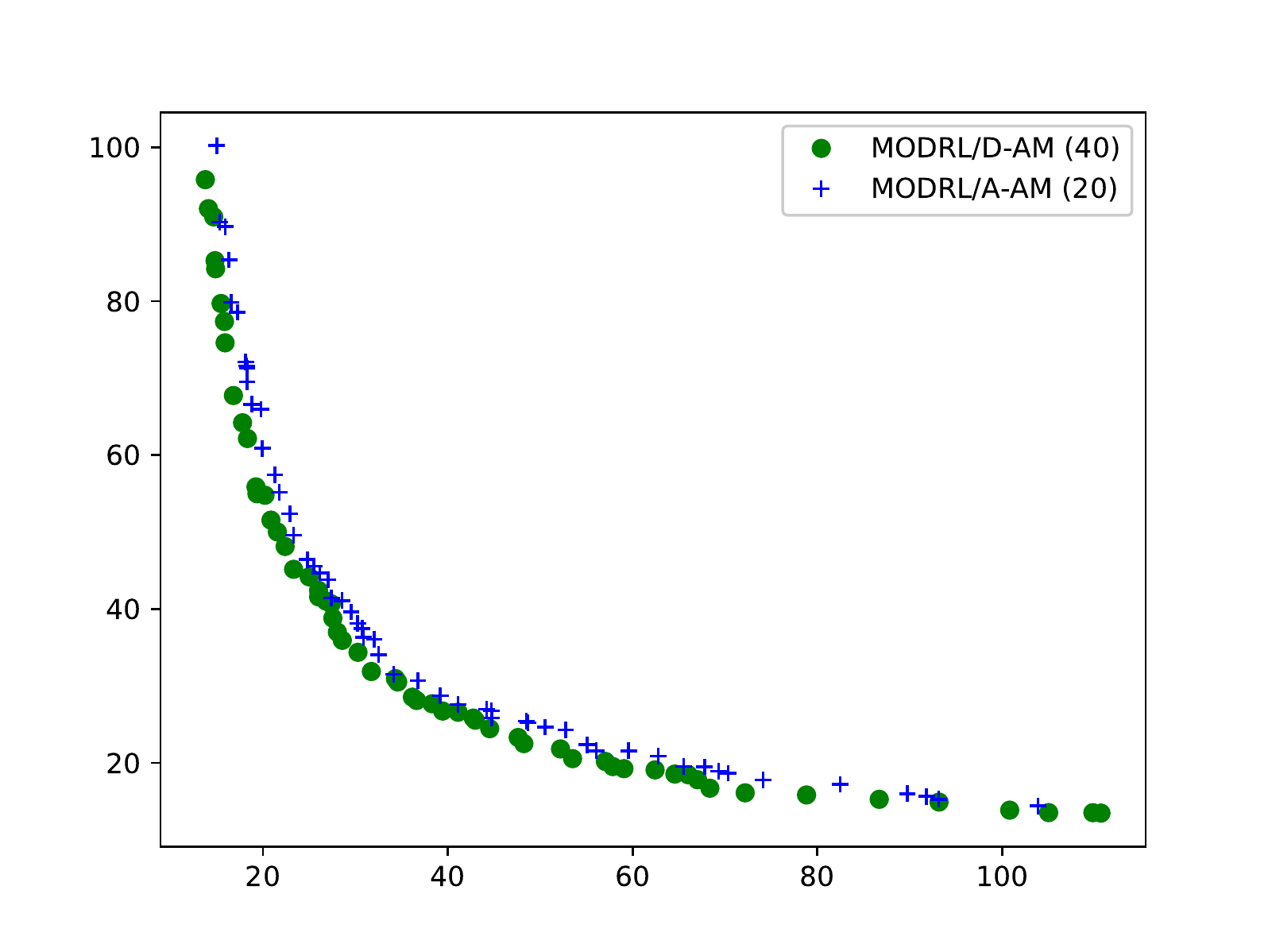}}
\caption{The PFs obtained by MODRL/D-AM (40) and MODRL/D-AM (20) in solving random instances with (a) 40 nodes, (b) 100 nodes, (c) 200 nodes}
\label{figure6}
\end{figure}

The result of the tested instances with different nodes is shown in Fig.~\ref{figure4}. By increasing the number of nodes, MODRL/D-AM (40) is able to get better performance in terms of convergence and diversity than that of DRL-MOA (40). Fig.~\ref{figure5} shows the performance of MODRL/D-AM (40) and DRL-MOA (40) on kroAB100, kroAB150 and kroAB200 instances. A significant improvement on convergence is observed for our method and the diversity achieved by our method is also slightly better.

Then, the performances of MODRL/D-AM (40) and MODRL/D-AM (20) are compared to investigate the influence of different number of nodes in training process. HV values in Table~\ref{table1} show that MODRL/D-AM (40) performs better on random instances with 40, 70, 100, 150 and 200 nodes than MODRL/D-AM (20). For the random instances with 20 nodes, MODRL/D-AM (40) performs similar to MODRL/D-AM (20), while MODRL/D-AM (40) is slightly worse. From the PFs obtained by MODRL/D-AM (40) and MODRL/D-AM (20) in Fig.~\ref{figure6}, a better performance is observed in terms of convergence and diversity. When training with instances with larger number of nodes, the model of MODRL/D-AM can learn to deal with more complex information about node features and structure features. Thus, a better model of MODRL/D-AM can be trained with more nodes.

From the experiment results above, it is observed that MODRL/D-AM has a good generalization performance in solving MOTSP. For MODRL/D-AM, the model trained with 40 nodes can be used to approximate the PF of problem instances with 200 nodes. In terms of convergence and diversity, MODRL/D-AM performs better than DRL-MOA.

The good performance of MODRL/D-AM indicates that the graph structure features are helpful in constructing solutions for MOTSP, and attention model can extract the structure information of a problem instance effectively. Thus, MODRL/D-AM can also be applied to other similar combinatorial optimization problems with graph structures such as multiobjective vehicle routing problem \cite{WJH1,WJH2}. Finally, there is still an issue that the solutions of MOTSP instances are not distributed evenly in our experiment, which needs further research.

\section{Conclusions}
This paper proposes an multiobjective deep reinforcement learning algorithm using decomposition and attention model. MODRL/D-AM adopts an attention model to model the subproblems of MOPs. The attention model can extract structure features as well as node features of problem instances. Thus, more useful structure information is used to generate better solutions. MODRL/D-AM is tested on MOTSP instances, and compared with DRL-MOA which uses pointer network to model the subproblems of MOTSP. The results show MODRL/D-AM achieves better performance. A good generalization performance on different size of problem instances is also observed for MODRL/D-AM.

\section*{Acknowledgement}
This work is supported by the National Key R\&D Program of China \\
(2018AAA0101203), and the National Natural Science Foundation of China (61673403, U1611262).

\bibliographystyle{splncs}
\bibliography{myreference}

\end{document}